\documentclass[acmsmall]{acmart}
\setcopyright{acmlicensed}
\copyrightyear{2025}
\acmYear{2025}
\acmDOI{10.1145/3737450}

\usepackage{color,colortbl}
\usepackage{todonotes} 
\usepackage{soul}
\usepackage{comment}
\usepackage{caption}
\usepackage{subcaption}
\usepackage{listings}

\usepackage[T1]{fontenc}
\usepackage[utf8]{inputenc}
\usepackage{arabtex}
\usepackage{utf8}
\setcode{utf8}

\usepackage{url}
\usepackage{soul}
\usepackage{xcolor} 
\usepackage{todonotes}
\definecolor{lightskyblue}{rgb}{0.53, 0.81, 0.98}

\usepackage{graphicx}
\definecolor{blue}{rgb}{0,0, 0.6}
\definecolor{dkgreen}{rgb}{0,0.6,0}
\definecolor{gray}{rgb}{0.5,0.5,0.5}
\definecolor{mauve}{rgb}{0.58,0,0.82}
\definecolor{mauve}{rgb}{0,0,0}
\definecolor{black}{rgb}{0,0,0}
\definecolor{tri}{rgb}{.25,.88,.82}
\definecolor{lilac}{rgb}{0.85,0.64,0.85}
\definecolor{lightblue}{rgb}{0.53, 0.81, 0.98}

\usepackage{tikz}
\usetikzlibrary{shapes.geometric, arrows.meta, positioning, fit, backgrounds}

\begin{document}


\title{Combating Misinformation in the Arab World: Challenges \& Opportunities}

\author{Azza Abouzied}
\email{azza@nyu.edu}
\orcid{}
\affiliation{%
  \institution{is an Associate Professor at New York University Abu Dhabi}
  \state{Abu Dhabi}
  \country{United Arab Emirates$^\ddagger$}
}

\author{Firoj Alam}
\email{fialam@hbku.edu.qa}
\affiliation{%
  \institution{is a Senior Scientist at Qatar Computing Research Institute}
  \state{Doha}
  \country{Qatar$^\ddagger$}
}

\author{Raian Ali}
\email{raali2@hbku.edu.qa}
\affiliation{%
  \institution{is a Professor at College of Science and Engineering, Hamad Bin Khalifa University}
  \state{Doha}
  \country{Qatar}
}

\author{Paolo Papotti}
\email{paolo.papotti@eurecom.fr}
\affiliation{%
  \institution{is a Professor at EURECOM}
  \state{Biot}
  \country{France$^\dagger$}
}

\begin{abstract}
Misinformation and disinformation
pose significant risks globally, with the Arab region facing unique vulnerabilities due to geopolitical instabilities, linguistic diversity, and cultural nuances. We explore these challenges through the key facets of combating misinformation: detection, tracking, mitigation and community-engagement. We shed light on how connecting with grass-roots fact-checking organizations, understanding cultural norms, promoting social correction, and creating strong collaborative information networks can create opportunities for a more resilient information ecosystem in the Arab world.
\end{abstract}


\newcommand{\azza}[1]{{{\color{orange} [Azza: #1]}}}
\newcommand{\eat}[1]{{}}
\maketitle

\renewcommand{\thefootnote}{\fnsymbol{footnote}}
\footnotetext[1]{Authors' names are listed in alphabetical order, with all authors contributing equally.}
\footnotetext[2]{Work done while visiting New York University Abu Dhabi.}
\footnotetext[3]{Corresponding authors.}
\renewcommand{\thefootnote}{\arabic{footnote}}

\section{Introduction}

Misinformation and disinformation are undoubtedly global risks, but the geopolitical instabilities and evolving crises make the Arab region particularly vulnerable. While misinformation includes false or misleading content, such as rumors, satire taken as fact, or conspiracy theories, disinformation is the intentional and targeted spread of such content to deceive or manipulate specific audiences~\cite{ndss}. To limit the spread and influence of misinformation, it is essential to advance research on technological methods for early detection, tracking, and mitigation, while also strengthening media literacy and promoting active citizen participation. 

At each stage in the fight against misinformation, we find challenges. In the Arab world, information is neither produced nor consumed in a single, universal language like Modern Standard Arabic (MSA). Instead, communication takes place across a wide range of dialects and languages --- including
Egyptian, Franco-Arabic, Gulf, Levantine --- as well as widely spoken languages like English and French.

This linguistic diversity adds significant complexity to the task of misinformation detection. To illustrate, the word \includegraphics[height=2.5ex]{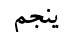}
in Tunisian Arabic (ynajjim) simply means ``he can'' or ``he is able'' while in Egyptian Arabic (yenaggim), it means ``he is practicing astrology,'' and is often used mockingly to imply that the person is fabricating claims, predicting an unknown future, or pretending to know things he does not.

Thus, the problem of misinformation detection can be understood as occurring within a multi-dimensional space, with each axis representing a key factor: dialectal variation, context (e.g., news articles, tweets, or social media posts), and modality (e.g., text, memes, videos, or images). In this high-dimensional space, the "curse of dimensionality'' becomes evident: there is simply not enough annotated data to train robust, automated detection tools. Compounding the challenge is the sparsity of authoritative information across these dimensions, which leaves the region particularly susceptible to disinformation tactics such as data void exploits. These arise when search engines or social media platforms return little to no credible content for a given query, creating an opening that malicious actors can exploit by flooding the space with misleading or false information.
Cultural and societal dynamics add further layers of complexity. Mistrust in formal media and fact-checking institutions, reliance on informal networks, such as family, tribal, or religious ties, limited media literacy, and low motivation all hinder participation in social correction efforts like Community Notes. Yet, despite these challenges, there remain meaningful opportunities for intervention and future research.

Our aim is to explore this complex research landscape by examining the technological methods surrounding misinformation in the Arab world. We make three contributions: First, we examine technological methods, models, and tools for automated detection, covering a wide range of tasks such as propaganda identification, check-worthiness estimation, and multimodal misinformation detection (Section~\ref{sec:early}). 
Second, we examine data void exploits as a form of disinformation attack and outline strategies for tracking and mitigating them (Section~\ref{sec:dv}).
Third, we investigate avenues for social correction, emphasizing the role of users and communities in countering misinformation, and highlight approaches that encourage community participation in mitigation efforts (Section~\ref{sec:correction}).

\section{Detection: Building AI systems}
\label{sec:early}

Datasets and benchmarks play a vital role in building automated misinformation detection methods. Unfortunately, there are few annotated datasets that cover the many dialectal variants of Arabic, the different modalities of communication such as images or videos, and the more nuanced misinformation detection tasks such as identifying propaganda markers, hate speech, or claim checkworthiness. The need for data is further driven by recent research that illustrates that data-hungry transformer-based, Arabic language models such as AraBERT and MARBERT consistently outperform traditional machine learning methods in misinformation detection tasks \cite{alotaibi2024review}.

Recent efforts are overcoming the data sparsity challenge by curating annotated text (AraNews \cite{elmadany2020machine}, AraFacts~\cite{sheikh-ali-etal-2021-arafacts}, and COVID-19-disinformation \cite{alam-etal-2021-fighting-covid}), 
and multimodal datasets (ArMeme \cite{alam-etal-2024-armeme}), as well as shared task benchmarks (CheckThat! Lab\footnote{\url{https://checkthat.gitlab.io}}, ArAIEval\footnote{\url{https://araieval.gitlab.io}}, and OSCAT\footnote{\url{https://osact5-lrec.github.io/}}). Even so, scaling these annotation and ground-truthing efforts is difficult: one not only has to find annotators that speak the different dialects, but also has to provide annotation guidelines\footnote{\url{https://www.digitqr.net/critical-digital-literacy/}} that reflect the linguistic and cultural norms of the annotators themselves. For example, an image depicting women in revealing attire may be considered inappropriate or offensive in certain Arabic cultures, highlighting the need for culturally sensitive annotation guidelines. 

In addition to research-driven efforts to create annotated datasets, grass-roots initiatives such as Misbar\footnote{\url{https://misbar.com}} and Fatabyyano\footnote{\url{https://fatabyyano.net}} where human fact-checkers identify misleading content, verify it, and provide evidence, can also support the work of creating rich repositories necessary for the further development and validation of automatic detection methods. Automated misinformation detection can in turn scale the efforts of human fact-checkers by bringing to their attention check worthy content. 
Bridging the gap between researchers and practitioners --- whether independent fact-checkers or news media agencies --- is a key opportunity for combating misinformation in the Arab world. Initiatives such as Tanbih\footnote{\url{https://tanbih.qcri.org/}} --- a research platform that provides the public with tools for detecting propaganda\footnote{Propaganda is the strategic use of true, or false information to promote a particular agenda, often blending facts with manipulation or emotionally-charged language to shape opinion.}, factuality, media bias, and framing --- represent promising initial efforts in this direction~\cite{zhang2019tanbih,hasanain-etal-2024-gpt}.

How to build robust and reliable automated misinformation detection tools remains an open problem. New techniques continuously emerge, often surpassing existing approaches. These include leveraging pretrained language models such as AraBERT and MARBERT \textit{vs.} classical deep learning architectures like BiLSTM and CNN, using general-purpose Large Language Models (LLMs) \textit{vs.} Arabic-only transformers, applying model-level fusion and attention mechanisms for multimodal data \textit{vs.} feature-level fusion, and incorporating metadata such as user engagement features \textit{vs.} relying solely on base content ~\cite{alotaibi2024review}. 

Recent benchmarks demonstrate considerable variability in state-of-the-art performance across tasks such as factuality assessment, propaganda detection, and claim verification, with accuracy scores ranging from 0.55 to 0.95~\cite{abdelali-etal-2024-larabench, kmainasi2024llamalensspecializedmultilingualllm,Yousef2024FakeNews}. Comparative evaluations between LLMs and task-specific models further suggest that, despite recent advances, there is still significant room for improvement in effectively addressing these tasks. In this rapidly evolving field, ongoing research and empirical validation are essential to ensure that emerging methods can be effectively adopted and applied to the persistent and context-specific challenges of detecting misinformation in the Arab world.

\section{Tracking and Mitigation: Combating Data Void Exploits}
\label{sec:dv}
Disinformation is misinformation’s motivated, coordinated, and targeted counterpart.
Disinformation is characterized by strategic agents, such as state-sponsored actors or regional interest groups, with access to resources and information dissemination assets, and a target victim demographic, whom they wish to influence~\cite{ndss}. By examining disinformation through a cybersecurity lens, we can better identify and categorize threats and vulnerabilities within the Arab region, and build tools to track such threats and effectively mitigate them through precise and cost-effective responses. 
One such threat the region to which is particularly susceptible to is a
\textit{data void exploit}.

A data void is a gap in the information ecosystem. A search begins with keywords or questions. When there is a dearth of information online that is relevant to the keywords, we are in a data void. Data voids are not inherently problematic. Random search strings lead us into voids. Disinformers, however, capitalize on the presence of data voids with respect to certain keywords or queries and the operation of search engines to drive information seekers to their narratives~\cite{boyd}. They fill the voids with their content before the emergence of authoritative information. Information seekers then self-discover the content by actively searching, deeming it authentic as it was ``found'' rather than passively shared with them. 
The exploitation of data voids can profoundly and permanently influence the beliefs of people. 

In the Arab region, data voids present unique challenges, especially amid frequent breaking news and emerging crises. Disinformers are quick to exploit these gaps during times of geopolitical instability, seizing the opportunity to spread fabricated narratives before credible sources can deliver verified information.

The linguistic diversity within the Arab region complicates efforts to track and mitigate disinformation. In this post-colonial context, communities communicate in multiple languages, including hybrids such as Franco-Arab, Arablish, and Arabizi. This complexity is further pronounced in cosmopolitan cities like Dubai and Doha, where diverse populations rely on a variety of global news sources and social media platforms. As each group maintains its affinity to a different linguistic or trusted news source, the lack of a universal authoritative informational source further exacerbates the potential for data voids and their exploits. 

Golebiewski and boyd argue that data void exploits are largely intractable without systematic, intentional and thoughtful management by the media and search platforms that host and index content~\cite{boyd}. Recently, as platforms struggle with balancing the individual right to free speech with society's need for trusted information, they are moving further away from the systematic, intentional and controlled management of disinformation~\cite{metagoesnotes}.

Despite the challenges posed by data voids, promising opportunities emerge for the Arab region. 
Our research into these exploits has led to the development of a language-agnostic tool capable of tracking the efficacy of an on-going exploit: we show that \textit{search result
rank} can determine the effectiveness and progress of both disinformation
or mitigation efforts with respect to a set of data void keywords in any language. We validated our tracking tool across both historical and contemporary data void case studies~\cite{datavoids}. By leveraging language models to automatically identify and label disinforming and mitigating narratives, we can effectively assess their influence in search queries.
We use adversarial game-theoretic simulations on a proxy model of the web to explore how mitigators can counter disinformers more effectively. In this setup, both sides act as competing agents with different resources, each trying to boost the visibility of their content in search results. Our results show that successful mitigation requires timely and strategic content placement. Crucially, we find that establishing high-influence information networks is core to a cost-effective response to an ongoing exploit by strategic disinformers. 
This amounts to creating a networked and coordinated coalition of fact-checking and credible, multilingual information sources. Therefore, even if search and media platforms were to disengage from disinformation management, it is possible for independent mitigators to build sufficient online information dissemination assets and boost their influence through linking them to each other to allow for an immediate and effective response to data void exploits.

\section{Community Engagement: Promoting Social Correction}
\label{sec:correction}

User correction relies on individuals flagging, debunking, and confronting those who post misinformation. Social correction, like crowd sourced fact-checking initiatives, e.g.,  Community Notes,
harness the collective intelligence of users to identify and rectify misinformation~\cite{0002TNDP22}. These collaborative efforts can be particularly effective in regions where diverse linguistic and cultural contexts require localized understanding. 
Despite its potential, people often hesitate to participate in such activities. Olson and LaPoe argue that the Spiral of Silence Theory applies in the context of user correction of misinformation on social media \cite{carter2018combating}; The theory
suggests that the reluctance to engage in acts requiring confrontation or opposing what appears to be the majority opinion stems from a human need to belong, leading to fear of isolation and then conformity.

The question then arises: why do people remain silent, and is there a misperception in how they reason about it? Reasons for avoiding the correction of misinformation can be grouped into four categories \cite{gurgun2024would}: relationship consequences (e.g., ``Will correcting others harm my relationship with them?''), negative impact on the person being challenged (e.g., ``Will they feel offended or seem less trustworthy?''), the perceived futility of the act (e.g., ``Is correcting misinformation even useful?''), and injunctive social norms (e.g., ``Is this behavior socially acceptable?''). Participants from two distinct cultural contexts, Arab and UK populations, exhibited misperceptions across these reasons. For instance, individuals overestimated the potential harm a correction might cause to relationships and the futility of challenging misinformation compared to reality. These misperceptions, while largely similar across cultures, significantly influence the willingness to engage in user correction~\cite{noman2023challenging}.

The identification of misperceptions creates an opportunity to leverage the social norms approach, which involves challenging individuals’ assumptions and encouraging behavioral change \cite{gurgun2024would}. This approach has been successfully applied in other domains, such as correcting misperceptions about smoking and alcohol consumption. It holds promise for addressing misperceptions about the utility and appropriateness of misinformation correction. While for it to be most effective it requires redesigning social media platforms, it can also be implemented through digital literacy campaigns and framing it as a pro-social act. Such a framing resonates with cultures that value group benefit, acts of donation, and altruism like the Arab culture. Digital nudging can also promote social correction but more research is required to determine which nudges are more impactful: Question stickers, for example that tag content with questions such as `is this source credible?' can be more effective than default comment boxes in encouraging social correction ~\cite{noman2024designing}.

A co-design study \cite{gurgun2024motivated} revealed that users prefer social media and online forums to be enhanced by fostering a sense of secure online environments, easing confrontations, facilitating access to reliable debunking information, and leveraging social recognition and social proof. However, translating this into actual designs that are both usable and effective remains a research challenge.

Attitudinal inoculation is a psychological technique used to build resistance to persuasion, including that found in misinformation posts and news. It works by exposing individuals to weak counterarguments or small doses of opposing views, triggering critical thinking and inoculating them, similar to a vaccine, by activating their defenses and preparing them to counter stronger future challenges to their beliefs \cite{Lewandowsky03072021}. Incorporating attitudinal inoculation into platform design has shown potential in enhancing resilience to persuasive online platforms, such as gambling \cite{cemiloglu2024explainability}. This inoculation-based strategy, combined with social norms messaging, could provide additional opportunities for behavior change if integrated into social media platforms to address misperceptions about the usefulness and relevance of the pro-social act of user correction. For example, it could occasionally prompt users to identify misleading and persuasive elements in a simulated post and define ways they can challenge majority silence and perceived social norms.
        
\section{Charting the Path Ahead}

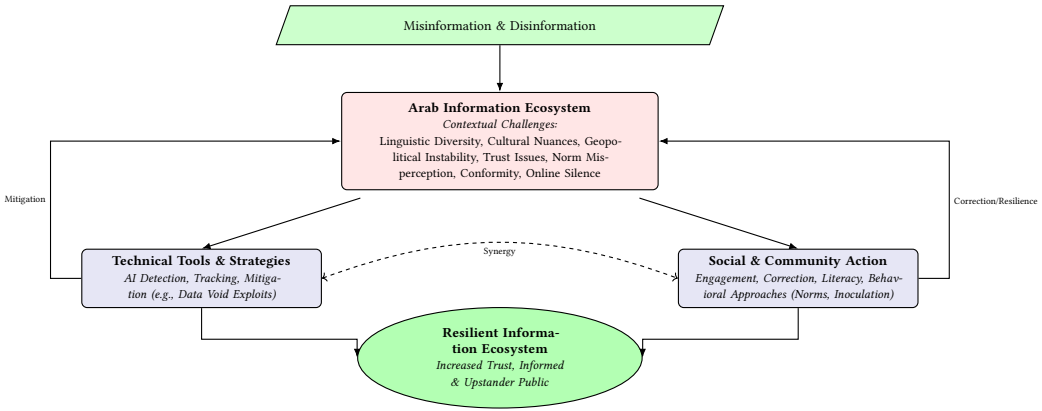
\begin{figure}[htbp]
\centering
\resizebox{\textwidth}{!}{
\begin{tikzpicture}[
    node distance=1cm and 1.5cm, 
    main/.style={rectangle, draw, thick, text centered, rounded corners, minimum height=1.5cm, text width=6cm, fill=blue!10},
    eco/.style={rectangle, draw, thick, text centered, rounded corners, minimum height=2.5cm, text width=8cm, fill=red!10},
    io/.style={trapezium, trapezium left angle=70, trapezium right angle=110, draw, thick, text centered, minimum height=1cm, fill=green!20, text width=5cm},
    goal/.style={ellipse, draw, thick, text centered, minimum height=1cm, fill=green!30, text width=5cm},
    arrow/.style={-Latex, thick}, 
    synergy/.style={<->, thick, dashed, bend angle=20} 
]

\node (misinfo) [io] {Misinformation \& Disinformation};

\node (ecosystem) [eco, below=1.2cm of misinfo] {
    \textbf{Arab Information Ecosystem} \\
    \textit{\small Contextual Challenges:} \\
    \small Linguistic Diversity, Cultural Nuances, Geopolitical Instability, Trust Issues, Norm Misperception, Conformity, Online Silence
};

\node (tech) [main, below left=1.5cm and 0.5cm of ecosystem] {
    \textbf{Technical Tools \& Strategies} \\
    \textit{\small AI Detection, Tracking, Mitigation (e.g., Data Void Exploits)}
};

\node (social) [main, below right=1.5cm and 0.5cm of ecosystem] {
    \textbf{Social \& Community Action} \\
    \textit{\small Engagement, Correction, Literacy, Behavioral Approaches (Norms, Inoculation)}
};

\node (goal) [goal, below=1.5cm of ecosystem, yshift=-1.5cm] { 
    \textbf{Resilient Information Ecosystem} \\
    \textit{\small Increased Trust, Informed \& Upstander Public}
};

\draw [arrow] (misinfo) -- (ecosystem);

\draw [arrow] (ecosystem.south west) ++(0.5, -0.2) -- (tech.north);
\draw [arrow] (ecosystem.south east) ++(-0.5, -0.2) -- (social.north);

\draw [arrow] (tech.west) -- ++(-0.8, 0) |- node[pos=0.25, above left, font=\scriptsize] {Mitigation} (ecosystem.west);
\draw [arrow] (social.east) -- ++(0.8, 0) |- node[pos=0.25, above right, font=\scriptsize] {Correction/Resilience} (ecosystem.east);

\draw [synergy] (tech.east) to[bend left] node[midway, below, font=\scriptsize] {Synergy} (social.west);

\draw [arrow] (tech.south) -- ++(0, -0.8) -| (goal.west);
\draw [arrow] (social.south) -- ++(0, -0.8) -| (goal.east);

\end{tikzpicture}
} 
\caption{A Socio-Technical Framework for Addressing Misinformation in the Arab World. Misinformation enters a complex ecosystem with distinct challenges. Technical tools and community efforts work together to reduce harm and encourage correction, helping to make the ecosystem more resilient.}
\label{fig:misinfo_framework_abstract}
\end{figure}

Effectively countering misinformation in the Arab region requires a coordinated socio-technical approach that brings together automated tools, networked mitigation strategies, and culturally grounded user engagement (See Figure \ref{fig:misinfo_framework_abstract}).
AI-powered automated systems are essential for detecting and flagging misleading content at scale. But these systems must be trained with sufficient regional data and guided by human oversight to ensure interventions are accurate, trusted, and contextually appropriate.

Beyond detection, our research shows that cost-effective responses to disinformation attacks such as data void exploits, require active monitoring and the development of high-influence networks of credible, multilingual sources. Linking these sources creates a resilient infrastructure that can respond rapidly, even when platforms step back from moderation. Fact-checking organizations, news media, and civil society can all invest in and amplify these networks.

Finally, engaging users in the correction process is critically important. In Arab societies, where public correction may be discouraged by social norms, social norm messaging (e.g., public campaigns that emphasize ``\textit{It's a good thing to correct misinformation online!}'') can help shift behavior from silence to action. Platforms can support this shift by designing features that frame correction as a positive, community-oriented, and altruistic act. Governments can amplify these strategies through targeted, paid campaigns on social media, especially during high-stakes moments of misinformation spread. Education and media literacy initiatives that psychologically inoculate communities are also effective, though they require long-term investment by public institutions. Together, these technical, institutional, and social efforts form a scalable, culturally attuned response to the evolving threat of misinformation.

\section*{Acknowledgments}
This publication was partly supported by NPRP 14 Cluster grant \# NPRP 14C-0916–210015 from the Qatar National Research Fund (a member of Qatar Foundation), the ASPIRE Award for Research Excellence (AARE-2020) grant AARE20-307, NYUAD CITIES through
Tamkeen - Research Institute Award CG001, and the ANR
project ATTENTION (ANR-21-CE23-0037).
The findings herein reflect the work and are solely the responsibility of the authors.
\bibliographystyle{ACM-Reference-Format}
\bibliography{bibliography}


\begin{thebibliography}{22}


\ifx \showCODEN    \undefined \def \showCODEN     #1{\unskip}     \fi
\ifx \showDOI      \undefined \def \showDOI       #1{#1}\fi
\ifx \showISBNx    \undefined \def \showISBNx     #1{\unskip}     \fi
\ifx \showISBNxiii \undefined \def \showISBNxiii  #1{\unskip}     \fi
\ifx \showISSN     \undefined \def \showISSN      #1{\unskip}     \fi
\ifx \showLCCN     \undefined \def \showLCCN      #1{\unskip}     \fi
\ifx \shownote     \undefined \def \shownote      #1{#1}          \fi
\ifx \showarticletitle \undefined \def \showarticletitle #1{#1}   \fi
\ifx \showURL      \undefined \def \showURL       {\relax}        \fi
\providecommand\bibfield[2]{#2}
\providecommand\bibinfo[2]{#2}
\providecommand\natexlab[1]{#1}
\providecommand\showeprint[2][]{arXiv:#2}

\bibitem[Abdelali et~al\mbox{.}(2024)]%
        {abdelali-etal-2024-larabench}
\bibfield{author}{\bibinfo{person}{Ahmed Abdelali}, \bibinfo{person}{Hamdy Mubarak}, \bibinfo{person}{Shammur Chowdhury}, \bibinfo{person}{Maram Hasanain}, \bibinfo{person}{Basel Mousi}, \bibinfo{person}{Sabri Boughorbel}, \bibinfo{person}{Samir Abdaljalil}, \bibinfo{person}{Yassine El~Kheir}, \bibinfo{person}{Daniel Izham}, \bibinfo{person}{Fahim Dalvi}, \bibinfo{person}{Majd Hawasly}, \bibinfo{person}{Nizi Nazar}, \bibinfo{person}{Youssef Elshahawy}, \bibinfo{person}{Ahmed Ali}, \bibinfo{person}{Nadir Durrani}, \bibinfo{person}{Natasa Milic-Frayling}, {and} \bibinfo{person}{Firoj Alam}.} \bibinfo{year}{2024}\natexlab{}.
\newblock \showarticletitle{{LA}ra{B}ench: Benchmarking {A}rabic {AI} with Large Language Models}. In \bibinfo{booktitle}{\emph{Proceedings of the 18th Conference of the European Chapter of the Association for Computational Linguistics (Volume 1: Long Papers)}}, \bibfield{editor}{\bibinfo{person}{Yvette Graham} {and} \bibinfo{person}{Matthew Purver}} (Eds.). \bibinfo{publisher}{Association for Computational Linguistics}, \bibinfo{address}{St. Julian{'}s, Malta}, \bibinfo{pages}{487--520}.
\newblock


\bibitem[Alam et~al\mbox{.}(2024)]%
        {alam-etal-2024-armeme}
\bibfield{author}{\bibinfo{person}{Firoj Alam}, \bibinfo{person}{Abul Hasnat}, \bibinfo{person}{Fatema Ahmad}, \bibinfo{person}{Md.~Arid Hasan}, {and} \bibinfo{person}{Maram Hasanain}.} \bibinfo{year}{2024}\natexlab{}.
\newblock \showarticletitle{{A}r{M}eme: Propagandistic Content in {A}rabic Memes}. In \bibinfo{booktitle}{\emph{Proceedings of the 2024 Conference on Empirical Methods in Natural Language Processing}}, \bibfield{editor}{\bibinfo{person}{Yaser Al-Onaizan}, \bibinfo{person}{Mohit Bansal}, {and} \bibinfo{person}{Yun-Nung Chen}} (Eds.). \bibinfo{publisher}{Association for Computational Linguistics}, \bibinfo{address}{Miami, Florida, USA}, \bibinfo{pages}{21071--21090}.
\newblock
\urldef\tempurl%
\url{https://doi.org/10.18653/v1/2024.emnlp-main.1173}
\showDOI{\tempurl}


\bibitem[Alam et~al\mbox{.}(2021)]%
        {alam-etal-2021-fighting-covid}
\bibfield{author}{\bibinfo{person}{Firoj Alam}, \bibinfo{person}{Shaden Shaar}, \bibinfo{person}{Fahim Dalvi}, \bibinfo{person}{Hassan Sajjad}, \bibinfo{person}{Alex Nikolov}, \bibinfo{person}{Hamdy Mubarak}, \bibinfo{person}{Giovanni Da~San~Martino}, \bibinfo{person}{Ahmed Abdelali}, \bibinfo{person}{Nadir Durrani}, \bibinfo{person}{Kareem Darwish}, \bibinfo{person}{Abdulaziz Al-Homaid}, \bibinfo{person}{Wajdi Zaghouani}, \bibinfo{person}{Tommaso Caselli}, \bibinfo{person}{Gijs Danoe}, \bibinfo{person}{Friso Stolk}, \bibinfo{person}{Britt Bruntink}, {and} \bibinfo{person}{Preslav Nakov}.} \bibinfo{year}{2021}\natexlab{}.
\newblock \showarticletitle{Fighting the {COVID}-19 Infodemic: Modeling the Perspective of Journalists, Fact-Checkers, Social Media Platforms, Policy Makers, and the Society}. In \bibinfo{booktitle}{\emph{Findings of the Association for Computational Linguistics: EMNLP 2021}}, \bibfield{editor}{\bibinfo{person}{Marie-Francine Moens}, \bibinfo{person}{Xuanjing Huang}, \bibinfo{person}{Lucia Specia}, {and} \bibinfo{person}{Scott Wen-tau Yih}} (Eds.). \bibinfo{publisher}{Association for Computational Linguistics}, \bibinfo{address}{Punta Cana, Dominican Republic}, \bibinfo{pages}{611--649}.
\newblock
\urldef\tempurl%
\url{https://doi.org/10.18653/v1/2021.findings-emnlp.56}
\showDOI{\tempurl}


\bibitem[Alotaibi and Al-Dossari(2024)]%
        {alotaibi2024review}
\bibfield{author}{\bibinfo{person}{Taghreed Alotaibi} {and} \bibinfo{person}{Hmood Al-Dossari}.} \bibinfo{year}{2024}\natexlab{}.
\newblock \showarticletitle{A Review of Fake News Detection Techniques for Arabic Language.}
\newblock \bibinfo{journal}{\emph{International Journal of Advanced Computer Science \& Applications}} \bibinfo{volume}{15}, \bibinfo{number}{1} (\bibinfo{year}{2024}).
\newblock


\bibitem[Carter~Olson and LaPoe(2018)]%
        {carter2018combating}
\bibfield{author}{\bibinfo{person}{Candi Carter~Olson} {and} \bibinfo{person}{Victoria LaPoe}.} \bibinfo{year}{2018}\natexlab{}.
\newblock \showarticletitle{Combating the digital spiral of silence: Academic activists versus social media trolls}.
\newblock \bibinfo{journal}{\emph{Mediating misogyny: Gender, technology, and harassment}} (\bibinfo{year}{2018}), \bibinfo{pages}{271--291}.
\newblock


\bibitem[Cemiloglu et~al\mbox{.}(2024)]%
        {cemiloglu2024explainability}
\bibfield{author}{\bibinfo{person}{Deniz Cemiloglu}, \bibinfo{person}{Selin Gurgun}, \bibinfo{person}{Emily Arden-Close}, \bibinfo{person}{Nan Jiang}, {and} \bibinfo{person}{Raian Ali}.} \bibinfo{year}{2024}\natexlab{}.
\newblock \showarticletitle{Explainability as a psychological inoculation: building resistance to digital persuasion in online gambling through explainable interfaces}.
\newblock \bibinfo{journal}{\emph{International Journal of Human--Computer Interaction}} \bibinfo{volume}{40}, \bibinfo{number}{23} (\bibinfo{year}{2024}), \bibinfo{pages}{8378--8396}.
\newblock


\bibitem[Elmadany et~al\mbox{.}(2020)]%
        {elmadany2020machine}
\bibfield{author}{\bibinfo{person}{Abdelrahim Elmadany}, \bibinfo{person}{Muhammad Abdul-Mageed}, \bibinfo{person}{Tariq Alhindi}, {et~al\mbox{.}}} \bibinfo{year}{2020}\natexlab{}.
\newblock \showarticletitle{Machine Generation and Detection of Arabic Manipulated and Fake News}. In \bibinfo{booktitle}{\emph{Proceedings of the Fifth Arabic Natural Language Processing Workshop}}. \bibinfo{pages}{69--84}.
\newblock


\bibitem[Golebiewski and boyd(2019)]%
        {boyd}
\bibfield{author}{\bibinfo{person}{Michael Golebiewski} {and} \bibinfo{person}{danah boyd}.} \bibinfo{year}{2019}\natexlab{}.
\newblock \bibinfo{title}{Data Voids: Where Missing Data Can Easily Be Exploited}.
\newblock
\newblock
\urldef\tempurl%
\url{https://datasociety.net/library/data-voids/}
\showURL{%
\tempurl}


\bibitem[Gurgun et~al\mbox{.}(2024a)]%
        {gurgun2024motivated}
\bibfield{author}{\bibinfo{person}{Selin Gurgun}, \bibinfo{person}{Emily Arden-Close}, \bibinfo{person}{Keith Phalp}, {and} \bibinfo{person}{Raian Ali}.} \bibinfo{year}{2024}\natexlab{a}.
\newblock \showarticletitle{Motivated by Design: A Codesign Study to Promote Challenging Misinformation on Social Media}.
\newblock \bibinfo{journal}{\emph{Human Behavior and Emerging Technologies}} \bibinfo{volume}{2024}, \bibinfo{number}{1} (\bibinfo{year}{2024}), \bibinfo{pages}{5595339}.
\newblock


\bibitem[Gurgun et~al\mbox{.}(2024b)]%
        {gurgun2024would}
\bibfield{author}{\bibinfo{person}{Selin Gurgun}, \bibinfo{person}{Muaadh Noman}, \bibinfo{person}{Emily Arden-Close}, \bibinfo{person}{Keith Phalp}, {and} \bibinfo{person}{Raian Ali}.} \bibinfo{year}{2024}\natexlab{b}.
\newblock \showarticletitle{How Would I Be Perceived If I Challenge Individuals Sharing Misinformation? Exploring Misperceptions in the UK and Arab Samples and the Potential for the Social Norms Approach}. In \bibinfo{booktitle}{\emph{International Conference on Persuasive Technology}}. Springer, \bibinfo{pages}{133--150}.
\newblock


\bibitem[Hasanain et~al\mbox{.}(2024)]%
        {hasanain-etal-2024-gpt}
\bibfield{author}{\bibinfo{person}{Maram Hasanain}, \bibinfo{person}{Fatema Ahmad}, {and} \bibinfo{person}{Firoj Alam}.} \bibinfo{year}{2024}\natexlab{}.
\newblock \showarticletitle{Can {GPT}-4 Identify Propaganda? Annotation and Detection of Propaganda Spans in News Articles}. In \bibinfo{booktitle}{\emph{Proceedings of the 2024 Joint International Conference on Computational Linguistics, Language Resources and Evaluation (LREC-COLING 2024)}}, \bibfield{editor}{\bibinfo{person}{Nicoletta Calzolari}, \bibinfo{person}{Min-Yen Kan}, \bibinfo{person}{Veronique Hoste}, \bibinfo{person}{Alessandro Lenci}, \bibinfo{person}{Sakriani Sakti}, {and} \bibinfo{person}{Nianwen Xue}} (Eds.). \bibinfo{publisher}{ELRA and ICCL}, \bibinfo{address}{Torino, Italia}, \bibinfo{pages}{2724--2744}.
\newblock
\urldef\tempurl%
\url{https://aclanthology.org/2024.lrec-main.244}
\showURL{%
\tempurl}


\bibitem[Isaac and Schleifer(2025)]%
        {metagoesnotes}
\bibfield{author}{\bibinfo{person}{Mike Isaac} {and} \bibinfo{person}{Theodore Schleifer}.} \bibinfo{year}{2025}\natexlab{}.
\newblock \showarticletitle{Meta Says It Will End Its Fact-Checking Program on Social Media Posts}.
\newblock \bibinfo{journal}{\emph{The New York Times}} (\bibinfo{year}{2025}).
\newblock
\urldef\tempurl%
\url{https://www.nytimes.com/2025/01/07/business/meta-community-notes-x.html}
\showURL{%
\tempurl}


\bibitem[Kmainasi et~al\mbox{.}(2025)]%
        {kmainasi2024llamalensspecializedmultilingualllm}
\bibfield{author}{\bibinfo{person}{Mohamed~Bayan Kmainasi}, \bibinfo{person}{Ali~Ezzat Shahroor}, \bibinfo{person}{Maram Hasanain}, \bibinfo{person}{Sahinur~Rahman Laskar}, \bibinfo{person}{Naeemul Hassan}, {and} \bibinfo{person}{Firoj Alam}.} \bibinfo{year}{2025}\natexlab{}.
\newblock \showarticletitle{{L}lama{L}ens: Specialized Multilingual {LLM} for Analyzing News and Social Media Content}. In \bibinfo{booktitle}{\emph{Findings of the Association for Computational Linguistics: NAACL 2025}}, \bibfield{editor}{\bibinfo{person}{Luis Chiruzzo}, \bibinfo{person}{Alan Ritter}, {and} \bibinfo{person}{Lu~Wang}} (Eds.). \bibinfo{publisher}{Association for Computational Linguistics}, \bibinfo{address}{Albuquerque, New Mexico}, \bibinfo{pages}{5627--5649}.
\newblock
\showISBNx{979-8-89176-195-7}
\urldef\tempurl%
\url{https://aclanthology.org/2025.findings-naacl.313/}
\showURL{%
\tempurl}


\bibitem[Lewandowsky and van~der Linden(2021)]%
        {Lewandowsky03072021}
\bibfield{author}{\bibinfo{person}{Stephan Lewandowsky} {and} \bibinfo{person}{Sander van~der Linden}.} \bibinfo{year}{2021}\natexlab{}.
\newblock \showarticletitle{Countering Misinformation and Fake News Through Inoculation and Prebunking}.
\newblock \bibinfo{journal}{\emph{European Review of Social Psychology}} \bibinfo{volume}{32}, \bibinfo{number}{2} (\bibinfo{year}{2021}), \bibinfo{pages}{348--384}.
\newblock
\urldef\tempurl%
\url{https://doi.org/10.1080/10463283.2021.1876983}
\showDOI{\tempurl}
\showeprint{https://doi.org/10.1080/10463283.2021.1876983}


\bibitem[Mannino et~al\mbox{.}(2024)]%
        {datavoids}
\bibfield{author}{\bibinfo{person}{Miro Mannino}, \bibinfo{person}{Junior Garcia}, \bibinfo{person}{Reem Hazim}, \bibinfo{person}{Azza Abouzied}, {and} \bibinfo{person}{Paolo Papotti}.} \bibinfo{year}{2024}\natexlab{}.
\newblock \showarticletitle{Data Void Exploits: Tracking {\&} Mitigation Strategies}. In \bibinfo{booktitle}{\emph{Proceedings of the 33rd {ACM} International Conference on Information and Knowledge Management, {CIKM} 2024, Boise, ID, USA, October 21-25, 2024}}, \bibfield{editor}{\bibinfo{person}{Edoardo Serra} {and} \bibinfo{person}{Francesca Spezzano}} (Eds.). \bibinfo{publisher}{{ACM}}, \bibinfo{pages}{1627--1637}.
\newblock
\urldef\tempurl%
\url{https://doi.org/10.1145/3627673.3679781}
\showDOI{\tempurl}


\bibitem[Mirza et~al\mbox{.}(2023)]%
        {ndss}
\bibfield{author}{\bibinfo{person}{Muhammad~Shujaat Mirza}, \bibinfo{person}{Labeeba Begum}, \bibinfo{person}{Liang Niu}, \bibinfo{person}{Sarah Pardo}, \bibinfo{person}{Azza Abouzied}, \bibinfo{person}{Paolo Papotti}, {and} \bibinfo{person}{Christina P{\"{o}}pper}.} \bibinfo{year}{2023}\natexlab{}.
\newblock \showarticletitle{Tactics, Threats {\&} Targets: Modeling Disinformation and its Mitigation}. In \bibinfo{booktitle}{\emph{30th Annual Network and Distributed System Security Symposium, {NDSS} 2023, San Diego, California, USA, February 27 - March 3, 2023}}. \bibinfo{publisher}{The Internet Society}.
\newblock
\urldef\tempurl%
\url{https://www.ndss-symposium.org/ndss-paper/tactics-threats-targets-modeling-disinformation-and-its-mitigation/}
\showURL{%
\tempurl}


\bibitem[Noman et~al\mbox{.}(2024)]%
        {noman2024designing}
\bibfield{author}{\bibinfo{person}{Muaadh Noman}, \bibinfo{person}{Selin Gurgun}, \bibinfo{person}{Keith Phalp}, {and} \bibinfo{person}{Raian Ali}.} \bibinfo{year}{2024}\natexlab{}.
\newblock \showarticletitle{Designing social media to foster user engagement in challenging misinformation: a cross-cultural comparison between the UK and Arab countries}.
\newblock \bibinfo{journal}{\emph{Humanities and Social Sciences Communications}} \bibinfo{volume}{11}, \bibinfo{number}{1} (\bibinfo{year}{2024}), \bibinfo{pages}{1--13}.
\newblock


\bibitem[Noman et~al\mbox{.}(2023)]%
        {noman2023challenging}
\bibfield{author}{\bibinfo{person}{Muaadh Noman}, \bibinfo{person}{Selin Gurgun}, \bibinfo{person}{Keith Phalp}, \bibinfo{person}{Preslav Nakov}, {and} \bibinfo{person}{Raian Ali}.} \bibinfo{year}{2023}\natexlab{}.
\newblock \showarticletitle{Challenging others when posting misinformation: a UK vs. Arab cross-cultural comparison on the perception of negative consequences and injunctive norms}.
\newblock \bibinfo{journal}{\emph{Behaviour \& Information Technology}} (\bibinfo{year}{2023}), \bibinfo{pages}{1--21}.
\newblock


\bibitem[Saeed et~al\mbox{.}(2022)]%
        {0002TNDP22}
\bibfield{author}{\bibinfo{person}{Mohammed Saeed}, \bibinfo{person}{Nicolas Traub}, \bibinfo{person}{Maelle Nicolas}, \bibinfo{person}{Gianluca Demartini}, {and} \bibinfo{person}{Paolo Papotti}.} \bibinfo{year}{2022}\natexlab{}.
\newblock \showarticletitle{Crowdsourced Fact-Checking at Twitter: How Does the Crowd Compare With Experts?}. In \bibinfo{booktitle}{\emph{Proceedings of the 31st {ACM} International Conference on Information {\&} Knowledge Management, Atlanta, GA, USA, October 17-21, 2022}}. \bibinfo{publisher}{{ACM}}, \bibinfo{pages}{1736--1746}.
\newblock
\urldef\tempurl%
\url{https://doi.org/10.1145/3511808.3557279}
\showDOI{\tempurl}


\bibitem[Sheikh~Ali et~al\mbox{.}(2021)]%
        {sheikh-ali-etal-2021-arafacts}
\bibfield{author}{\bibinfo{person}{Zien Sheikh~Ali}, \bibinfo{person}{Watheq Mansour}, \bibinfo{person}{Tamer Elsayed}, {and} \bibinfo{person}{Abdulaziz Al{-}Ali}.} \bibinfo{year}{2021}\natexlab{}.
\newblock \showarticletitle{{A}ra{F}acts: The First Large {A}rabic Dataset of Naturally Occurring Claims}. In \bibinfo{booktitle}{\emph{Proceedings of the Sixth Arabic Natural Language Processing Workshop}}, \bibfield{editor}{\bibinfo{person}{Nizar Habash}, \bibinfo{person}{Houda Bouamor}, \bibinfo{person}{Hazem Hajj}, \bibinfo{person}{Walid Magdy}, \bibinfo{person}{Wajdi Zaghouani}, \bibinfo{person}{Fethi Bougares}, \bibinfo{person}{Nadi Tomeh}, \bibinfo{person}{Ibrahim Abu~Farha}, {and} \bibinfo{person}{Samia Touileb}} (Eds.). \bibinfo{publisher}{Association for Computational Linguistics}, \bibinfo{address}{Kyiv, Ukraine (Virtual)}, \bibinfo{pages}{231--236}.
\newblock
\urldef\tempurl%
\url{https://aclanthology.org/2021.wanlp-1.26/}
\showURL{%
\tempurl}


\bibitem[Yousef et~al\mbox{.}(2024)]%
        {Yousef2024FakeNews}
\bibfield{author}{\bibinfo{person}{Mohammed~Abbas Yousef}, \bibinfo{person}{Abeer ElKorany}, {and} \bibinfo{person}{Hanaa Bayomi}.} \bibinfo{year}{2024}\natexlab{}.
\newblock \showarticletitle{Fake-news detection: a survey of evaluation Arabic datasets}.
\newblock \bibinfo{journal}{\emph{Social Network Analysis and Mining}} \bibinfo{volume}{14}, \bibinfo{number}{1} (\bibinfo{year}{2024}), \bibinfo{pages}{225}.
\newblock
\showISSN{1869-5469}
\urldef\tempurl%
\url{https://doi.org/10.1007/s13278-024-01370-2}
\showDOI{\tempurl}


\bibitem[Zhang et~al\mbox{.}(2019)]%
        {zhang2019tanbih}
\bibfield{author}{\bibinfo{person}{Yifan Zhang}, \bibinfo{person}{Giovanni Da~San~Martino}, \bibinfo{person}{Alberto Barr{\'o}n-Cede{\~n}o}, \bibinfo{person}{Salvatore Romeo}, \bibinfo{person}{Jisun An}, \bibinfo{person}{Haewoon Kwak}, \bibinfo{person}{Todor Staykovski}, \bibinfo{person}{Israa Jaradat}, \bibinfo{person}{Georgi Karadzhov}, \bibinfo{person}{Ramy Baly}, {et~al\mbox{.}}} \bibinfo{year}{2019}\natexlab{}.
\newblock \showarticletitle{Tanbih: Get To Know What You Are Reading}. In \bibinfo{booktitle}{\emph{Proceedings of the 2019 Conference on Empirical Methods in Natural Language Processing and the 9th International Joint Conference on Natural Language Processing (EMNLP-IJCNLP): System Demonstrations}}. \bibinfo{pages}{223--228}.
\newblock


\end{thebibliography}

\end{document}